\documentclass[a4paper,fleqn,review,longmktitle]{cas-sc}
\usepackage[numbers,sort&compress]{natbib}
\usepackage{tgtermes}
\usepackage{tgheros}
\usepackage{tgcursor}
\stmausetup{size=\fontsize{8}{10}\selectfont,align=\centering,color=black}
\stmaffsetup{size=\fontsize{8}{10}\selectfont,shape=\itshape,
  align=\centering,color=black,before=0pt,weight=\mdseries}
\newcommand{\PRcaptionfont}{\rmfamily\fontsize{8}{10}\selectfont}
\makeatletter
\let\PRstandardcaption\@makecaption
\long\def\@makecaption#1#2{%
  \begingroup\PRcaptionfont\PRstandardcaption{#1}{#2}\endgroup}
\makeatother
\ExplSyntaxOn
\cs_if_exist:NF \vbox_unpack_clear:N
  {\cs_new_eq:NN\vbox_unpack_clear:N\vbox_unpack_drop:N}
\cs_new_eq:NN \PRoriginalFigureCaption \__make_fig_caption:nn
\cs_set_protected:Npn \__make_fig_caption:nn #1#2
  {\group_begin:\cs_set_eq:NN\small\PRcaptionfont
   \PRoriginalFigureCaption{#1}{#2}\group_end:}
\cs_new_eq:NN \PRoriginalTableCaption \__make_tbl_caption:nn
\cs_set_protected:Npn \__make_tbl_caption:nn #1#2
  {\group_begin:\cs_set_eq:NN\small\PRcaptionfont
   \PRoriginalTableCaption{#1}{#2}\group_end:}
\ExplSyntaxOff
\RenewDocumentEnvironment{Abstract}{o}{%
  \par\begingroup\parindent=0pt
  \noindent\textbf{Abstract}\par\medskip
  \normalsize\doublespacing\ignorespaces
}{\par\endgroup\medskip\copy\csname g_stm_key_box\endcsname}
\RenewDocumentEnvironment{keywords}{O{Keywords}}{%
  \global\setbox\csname g_stm_key_box\endcsname=\vtop\bgroup
  \hsize=\textwidth\parindent=0pt\normalsize\doublespacing
  \def\sep{; }\noindent\textit{#1}: \ignorespaces
}{\par\egroup}
\AtBeginEnvironment{tabular}{\rmfamily\normalsize}
\AtBeginEnvironment{tabular*}{\rmfamily\normalsize}
\AtBeginEnvironment{tabularx}{\rmfamily\normalsize}
\ExplSyntaxOn
\cs_set:Npn \__first_footerline:
  { {\rmfamily\small\itshape Preprint~submitted~to~Pattern~Recognition} }
\cs_new_protected:Npn \PRcontactparagraph
  {
    \leftskip=0pt\rightskip=0pt
    \parfillskip=0pt~plus~1fil
    \spaceskip=0pt\xspaceskip=0pt\parindent=0pt
    \emergencystretch=1em
  }
\RenewDocumentCommand \emailauthor { m m }
  {
    \int_gincr:N \g_ead_int
    \seq_gput_right:Nn \g_stm_ead_seq
      {
        {\urlstyle{same}
         \def\UrlBreaks{\do\@\do\.\do\-\do\_}
         \nolinkurl{#1}}
        \parsename{#2}\space(\eadauthor)
      }
  }
\RenewDocumentCommand \printemails { }
  {
    \group_begin:
    \int_compare:nNnT {\g_ead_int} > {0}
      {
        \tex_let:D \thefootnote \relax
        \footnotetext
          {
            \PRcontactparagraph
            \textit{Email~addresses:}\c_space_token
            \seq_use:Nn \g_stm_ead_seq {;\c_space_token}
            \par
          }
      }
    \group_end:
  }
\RenewDocumentCommand \orcidauthor { m m }
  {
    \seq_gput_right:Nn \g_stm_orcid_seq
      {
        {\urlstyle{same}\ttfamily
         \def\UrlBreaks{\do\-}\nolinkurl{#1}}
        \parsename{#2}\mbox{\space(\eadauthor)}
      }
  }
\RenewDocumentCommand \printorcid { }
  {
    \group_begin:
    \seq_if_empty:NF \g_stm_orcid_seq
      {
        \tex_let:D \thefootnote \relax
        \footnotetext
          {
            \PRcontactparagraph
            \mbox{\textsc{orcid}(s):\space}
            \seq_use:Nn \g_stm_orcid_seq {;\c_space_token}
            \par
          }
      }
    \group_end:
  }
\ExplSyntaxOff
\AtBeginDocument{}
\makeatletter
\patchcmd{\section}{15pt}{10pt}{}{\PackageWarning{PRlayout}{Section spacing patch failed}}
\patchcmd{\subsection}{10pt}{6pt}{}{\PackageWarning{PRlayout}{Subsection spacing patch failed}}
\patchcmd{\subsubsection}{10pt}{6pt}{}{\PackageWarning{PRlayout}{Subsubsection spacing patch failed}}
\apptocmd{\NAT@bibsetnum}{%
  \setlength{\itemindent}{\leftmargin}%
  \setlength{\leftmargin}{0pt}%
}{}{\PackageWarning{PRlayout}{Bibliography indentation patch failed}}
\makeatother
\usepackage{placeins}
\usepackage{rotating}

\newif\ifCLASSINFOpdf
\CLASSINFOpdftrue
\newif\ifCLASSOPTIONcaptionsoff
\CLASSOPTIONcaptionsofffalse
\usepackage{float}
\usepackage{tabularx}
\newcommand{\secondbest}[1]{%
  \begingroup\setbox0=\hbox{#1}%
  \leavevmode\rlap{\raisebox{-0.45ex}{\rule{\wd0}{0.35pt}}}\box0\endgroup}

\ifCLASSINFOpdf
\else
\fi
\begin{document}
	\let\WriteBookmarks\relax
	\def\floatpagepagefraction{1}
	\def\textpagefraction{.001}

	\shorttitle{}
	\shortauthors{Gong et al.}
		%
\title[mode=title,size=\fontsize{14}{16}\selectfont,align=\centering,color=black,weight=\mdseries]{Next-Frequency Autoregressive Modeling for Remote Sensing Change Detection}
		%
		%
		%
		
		\author[1,2]{Luqi Gong}[orcid=0009-0000-8744-8630]
		\ead{luqi@zhejianglab.org}
		\ead{luqi@bupt.edu.cn}
		
		\author[1]{Yue Chen}[orcid=0009-0008-4975-4854]
		\ead{chenyue@zhejianglab.org}
		
		\author[1]{Qixin Xie}[orcid=0009-0002-4457-966X]
		\ead{keshawn.hsieh@zhejianglab.org}
		
		\author[1]{Ziqiang Chen}[orcid=0009-0004-5138-9893]
		\ead{chenzq@zhejianglab.org}
		
		\author[1]{Chao Li}[orcid=0000-0002-5343-1862]
		\cormark[1]
		\ead{lichao@zhejianglab.org}
		
		\author[3]{Xuefeng Zhao}[orcid=0000-0002-4696-6228]
		\cormark[1]
		\ead{xuefengzhao@scau.edu.cn}
		
		\affiliation[1]{organization={Research Center for Space Computing System, Zhejiang Lab},
			addressline={Kechuang Avenue, Zhongtai Sub-District, Yuhang District},
			city={Hangzhou},
			postcode={311121},
			state={Zhejiang},
			country={China}}
		\affiliation[2]{organization={State Key Laboratory of Networking and Switching Technology, Beijing University of Posts and Telecommunications},
			addressline={10 Xitucheng Road, Haidian District},
			city={Beijing},
			postcode={100876},
			country={China}}
		\affiliation[3]{organization={College of Economics and Management, South China Agricultural University},
			addressline={483 Wushan Road, Tianhe District},
			city={Guangzhou},
			postcode={510642},
			country={China}}
		\cortext[1]{Corresponding author.}

	\begin{abstract}
		Remote sensing change detection aims to identify land-cover changes from bi-temporal images. Most existing methods follow a one-shot dense prediction paradigm, directly regressing a change mask from fused features. However, such approaches overlook the intrinsic frequency characteristics of change patterns. To address these limitations, we propose Freq-RemoteVAR, a frequency autoregressive framework that reformulates change detection as a structured generation problem in the frequency domain. Instead of predicting the change mask in a single step, we introduce a next-frequency prediction paradigm, where change information is progressively generated from coarse to fine. 
		To support this formulation, we design a frequency-aware mask tokenization strategy that decomposes change supervision into multi-frequency token targets via Fourier transformation and quantization. Based on this representation, we develop a Frequency VAR Transformer, which performs causal autoregressive modeling over frequency tokens. The model starts from learned mask queries and progressively predicts frequency-level tokens conditioned on previously generated tokens and bi-temporal image features, effectively capturing long-range dependencies across frequency scales. Furthermore, we introduce Scale-Aligned RoPE Cross Attention (SRCA) module, which aligns frequency-domain mask queries with spatial-domain bi-temporal features under a unified coordinate system, enhancing spatial-frequency consistency during generation. To mitigate the impact of unreliable signals, we propose a Change-quality Control module that adaptively modulates the generation process through dynamic normalization, attention biasing, and spatial offset adjustment, thereby suppressing pseudo-change responses and improving robustness. Extensive experiments on CDD, GZ-CD, and LEVIR-CD demonstrate that Freq-RemoteVAR consistently outperforms existing methods, particularly in challenging scenarios with complex appearance variations and noisy disturbances.
		
	\end{abstract}

	

	\maketitle
\setlength{\abovedisplayskip}{2pt plus 2pt minus 1pt}
\setlength{\belowdisplayskip}{2pt plus 2pt minus 1pt}
\setlength{\abovedisplayshortskip}{2pt plus 2pt}
\setlength{\belowdisplayshortskip}{2pt plus 2pt minus 1pt}
\setlength{\textfloatsep}{6pt plus 2pt minus 1pt}
\setlength{\floatsep}{8pt plus 2pt minus 1pt}
\setlength{\intextsep}{6pt plus 2pt minus 1pt}
\brokenpenalty=100

	%

	\section{Introduction}
	
	Remote sensing change detection (RSCD) aims to identify land-cover changes from multi-temporal remote sensing images acquired over the same geographical area. It plays a fundamental role in urban expansion monitoring, disaster assessment, land-use analysis, and environmental surveillance~\cite{radke2005image,hussain2013change}. With the increasing availability of high-resolution remote sensing images, change detection models can access richer textures, edges, and geometric structures. However, bi-temporal images are often affected by illumination variation, shadows, seasonal changes, sensor noise, and slight misregistration~\cite{daudt2018fully,bai2023deep}. These non-semantic discrepancies may introduce numerous pseudo changes, making it challenging to distinguish genuine land-cover changes from imaging disturbances.
	
	\begin{figure}[pos=!t,abovecap=6pt,belowcap=3pt]
		\centering
		\includegraphics[width=0.97\linewidth]{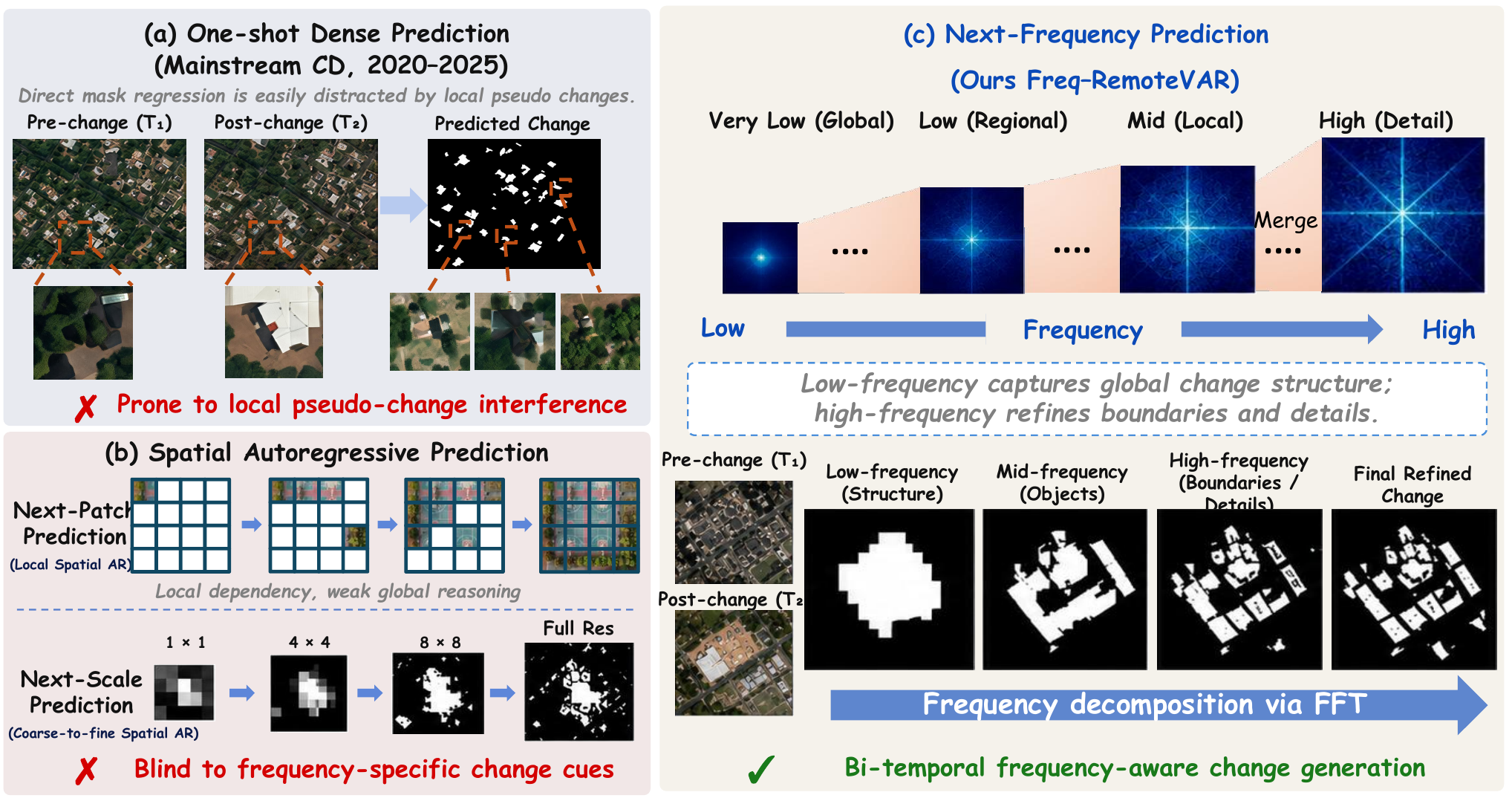}
		\caption{Motivation of the proposed Freq-RemoteVAR. 
			(a) Mainstream one-shot dense prediction methods directly regress the final change mask from bi-temporal images, which are easily distracted by local pseudo changes such as shadows, texture variations, and slight misregistration. 
			(b) Spatial autoregressive prediction generates change maps in a patch-wise or scale-wise manner, but it mainly models spatial progression and remains insensitive to frequency-specific change cues. 
			(c) Our Freq-RemoteVAR reformulates remote sensing change detection as next-frequency prediction. 
			By generating change masks from low to high frequency under bi-temporal guidance, our method enables more robust and structured change modeling.} 
		\label{fig:motivation}
	\end{figure}
	
	Recent deep learning methods have significantly advanced RSCD. CNN-based approaches leverage Siamese architectures, multi-scale fusion, and attention mechanisms to capture local textures and boundaries~\cite{chen2020spatial,fang2021snunet,zhang2024adaptive}, while Transformer-based methods improve global context modeling through self-attention~\cite{bandara2022transformer,zhang2022swinsunet,chen2022remote}. More recently, state space models and Mamba-based architectures have been introduced to efficiently model long-range dependencies~\cite{gu2023mamba,chen2024changemamba}. Despite their architectural differences, many methods produce the final change mask through one-shot dense prediction, as illustrated in Fig.~\ref{fig:motivation}(a). Their hierarchical features can encode both regional context and local details, but the output formulation does not explicitly condition the prediction of fine spatial variations on previously predicted change structures. This motivates a complementary formulation in which regional structure and local detail are connected through a progressive prediction process. Such a dependency is particularly relevant when local differences alone provide insufficient evidence for distinguishing genuine changes from appearance disturbances~\cite{saha2019unsupervised,peng2019end}. 
	
	Visual autoregressive (VAR) modeling offers a new perspective for change detection. Instead of directly predicting dense outputs, autoregressive models generate target representations progressively in a token space~\cite{van2016pixel}. Recent Visual Autoregressive models adopt a next-scale prediction strategy, generating images from low-resolution scales to high-resolution scales and achieving better efficiency than traditional next-token generation~\cite{tian2024visual}. Inspired by this idea, several change detection methods~\cite{korkmaz2026remotevar} formulate change map prediction as a progressive generation process. As illustrated in Fig.~\ref{fig:motivation}(b), such spatial autoregressive prediction organizes the generation process according to token-map resolution. 
	
	In contrast, we observe that the structural composition of a change mask can be naturally organized in the frequency domain. As illustrated in Fig.~\ref{fig:motivation}(c), low-frequency components mainly determine the approximate extent and connectivity of changed regions, mid-frequency components describe object-level shapes, and high-frequency components encode boundaries, small structures, and local details. This frequency hierarchy provides a structure-to-detail organization of change information. Importantly, it differs from spatial-scale autoregression: spatial-scale prediction changes the sampling resolution of token maps, whereas frequency-autoregressive prediction explicitly changes the frequency content represented at each generation stage. We therefore formulate frequency progression as conditional generation, in which previously generated frequency groups provide structural context and bi-temporal image features provide observational evidence for subsequent predictions. As illustrated in Fig.~\ref{fig:motivation}(c), this formulation allows finer variations to be inferred in the context of both preceding predictions and the input image pair.
	
	Based on this formulation, we propose Freq-RemoteVAR, a frequency-autoregressive framework for remote sensing change detection. Different from existing methods that use frequency information merely for feature enhancement, we directly adopt the frequency hierarchy of the target change mask as the generation order. To effectively exploit bi-temporal remote sensing images, we construct a bi-temporal fused memory by integrating pre-change features, post-change features, absolute difference features, and correlation features. This memory serves as conditional image evidence during frequency-token generation. Furthermore, we design a Frequency VAR Transformer to model causal dependencies among frequency tokens and query reliable change evidence from the bi-temporal memory.
    To align frequency tokens with multi-scale bi-temporal features, we introduce Scale-Aligned RoPE Cross-Attention (SRCA), which maps mask queries and bi-temporal memory tokens into a unified image-space coordinate system and performs scale-aligned rotary positional interaction. This design strengthens spatial correspondence across different frequency levels and feature scales, particularly in the presence of slight local displacement. To suppress the propagation of pseudo changes, we further develop Change-Quality Control (CQC), which dynamically estimates change reliability from bi-temporal features and modulates the autoregressive generation process.
	
	The main contributions of this paper are summarized as follows:
    \begin{itemize}
        \item We propose a frequency-autoregressive paradigm for remote sensing change detection, reformulating RSCD from one-shot dense prediction into low-to-high frequency change-mask generation.
        
        \item We develop a Frequency VAR Transformer, where bi-temporal fused memory provides reliable image evidence for progressive frequency-token generation.
        
        \item We introduce Scale-Aligned RoPE Cross-Attention to align frequency-mask queries with multi-scale bi-temporal features in a unified image-space coordinate system, thereby strengthening spatial correspondence across frequency levels and feature scales.
        
        \item We propose Change-Quality Control to suppress local pseudo-change propagation and improve boundary stability in complex remote sensing scenes.
    \end{itemize}
	
	\section{Related Works}
	
	\subsection{Deep Learning-based Remote Sensing Change Detection}
	
	Deep learning has substantially advanced remote sensing change detection (RSCD) by enabling end-to-end learning of discriminative representations from bi-temporal images. Early deep learning methods were mainly based on convolutional neural networks (CNNs), where Siamese architectures became the dominant framework due to their ability to extract comparable features from two temporal observations using shared parameters. Representative methods such as DSIFN~\cite{zhang2020dsifn}, SNUNet-CD~\cite{fang2021snunet}, DASNet~\cite{chen2020dasnet}, and STADE-CDNet~\cite{li2024stade} introduced deeply supervised fusion, dense feature interaction, and attention mechanisms to improve change localization and boundary delineation. Subsequent studies further incorporated multi-scale feature aggregation, object-level constraints, and contextual modeling strategies to enhance the detection of subtle changes and small objects~\cite{liu2021dtcdscn,li2022change}. Benefiting from self-attention mechanisms, Transformers can model long-range interactions and global contextual information more effectively than conventional CNNs. BIT~\cite{chen2022remote} introduced Transformer-based token interaction for bi-temporal feature modeling and demonstrated strong performance on multiple RSCD benchmarks. ChangeFormer~\cite{bandara2022transformer} further leveraged hierarchical Transformer representations and multi-scale decoding to improve semantic consistency across different spatial resolutions. More recent studies~\cite{li2022transunetcd,pan2024m} have explored hybrid CNN-Transformer architectures that combine local feature extraction with global context modeling. Despite their effectiveness, Transformer-based methods generally suffer from high computational complexity when processing high-resolution remote sensing imagery.
	
	Recently, state-space models (SSMs) and Mamba-based architectures have emerged as efficient alternatives for long-range visual modeling. By employing selective scanning mechanisms, these models can capture global dependencies with linear computational complexity. VMamba~\cite{liu2024vmamba} demonstrated the effectiveness of visual state-space modeling for dense prediction tasks. Building upon this idea, several RSCD methods introduced temporal interaction modules, cross-temporal scanning strategies, and global-local fusion mechanisms to improve change representation~\cite{zhao2025st,wang2025bi}. For example, GLMamba~\cite{liu2026glmamba} combines global contextual modeling and local detail preservation within a unified state-space framework, achieving competitive performance while maintaining computational efficiency.

	\subsection{Frequency-domain Modeling for Change Detection}
	
	Recent studies have explored frequency-domain representations for RSCD. Fourier-based approaches transform spatial features into the spectral domain and perform frequency-aware feature interaction to suppress irrelevant variations caused by illumination changes or atmospheric conditions~\cite{zhang2025change}. FSG-Net~\cite{xie2026fsg} utilizes frequency-spatial guidance based on wavelet decomposition to suppress pseudo changes and improve boundary localization. FIMP~\cite{chen2024high} employed the Fourier transform to adaptively filter bitemporal features in the frequency domain while mining the optimized bitemporal features relevant to the change detection task. These studies demonstrate that frequency-domain representations can effectively improve robustness against noise, seasonal variations, and slight misregistration. Beyond RSCD, frequency-domain modeling has also shown effectiveness in semantic segmentation, image restoration, and visual recognition tasks, such as frequency-channel attention, Fourier convolution, global frequency filtering, and frequency-domain reconstruction losses~\cite{qin2021fcanet,chi2020fast,rao2021global,jiang2021focal,yao2022wave,nie2024wavelet}.
	
	
	\subsection{Visual Autoregressive Modeling}
	
	Recently, Visual AutoRegressive modeling (VAR)~\cite{tian2024visual} reformulated image generation as a next-scale prediction problem, where token maps are progressively generated from coarse resolutions to finer resolutions. Compared with conventional next-token prediction, next-scale autoregression better preserves spatial structures and naturally captures hierarchical visual representations. Autoregressive paradigms have also begun to emerge in dense prediction tasks. Instead of directly regressing outputs in a single forward pass, these methods formulate segmentation or structured prediction as a progressive token generation process, enabling explicit modeling of global-to-local dependencies~\cite{zheng2026seg,jiao2026flexvar}. This line of research is closely related to discrete visual representation learning, where VQ-VAE~\cite{van2017neural}, VQGAN~\cite{esser2021taming}, DALL-E~\cite{ramesh2021zero}, and MaskGIT~\cite{chang2022maskgit} demonstrate that visual contents can be represented and generated in a discrete token space. In remote sensing change detection, RemoteVAR~\cite{korkmaz2026remotevar} is among the first attempts to introduce visual autoregressive modeling into change mask prediction. It discretizes change masks into latent tokens and progressively generates change representations under the guidance of bi-temporal image features. However, existing autoregressive RSCD methods generally organize generation according to spatial scales, following a coarse-to-fine resolution hierarchy. While effective, this strategy does not explicitly exploit the frequency characteristics of change masks. 
	
	Different from previous spatial-scale autoregressive frameworks, our method introduces a frequency-aware autoregressive paradigm for RSCD. Instead of progressively generating masks across resolutions, the proposed framework predicts change representations from low-frequency semantics to high-frequency details. This design explicitly aligns the generation order with the hierarchical frequency composition of change masks, enabling the model to first establish global change regions and subsequently refine object structures and boundaries. 

	\FloatBarrier
	\section{Methodology}
    
    Throughout this section, $I_1$ and $I_2$ denote the pre-change and post-change images, respectively. The ground-truth and predicted change masks are denoted by $M^{gt}$ and $\hat{M}$, while $\mathcal{M}$ represents the bi-temporal fused memory. The projected condition tokens are denoted by \(C=\phi_c(\mathcal{M})\). We use $T_k$ to denote the discrete token-index map of the $k$-th frequency band and $E_k$ to denote its corresponding codebook embedding map.

\begin{figure}[pos=htp,abovecap=6pt,belowcap=3pt]
		\centering
		\includegraphics[width=0.97\linewidth]{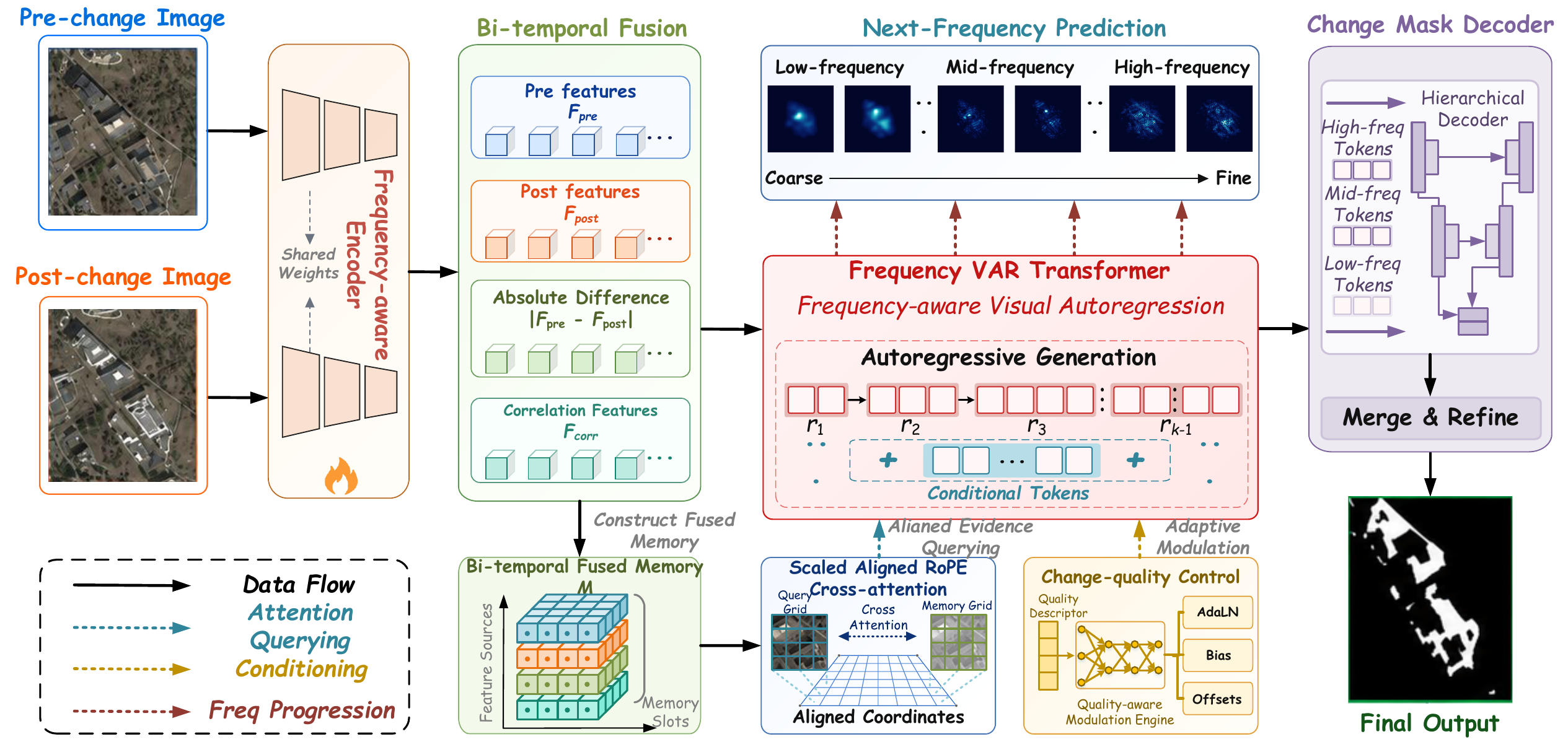}
		\caption{Overall framework of Freq-RemoteVAR. 
			The pre- and post-change images are first encoded into multi-scale bi-temporal features and fused into a conditional memory. 
			Starting from learned mask queries, the Frequency VAR Transformer progressively predicts frequency change tokens under the guidance of Scale-Aligned RoPE Cross Attention and Change-quality Control. 
			The generated tokens are finally decoded and refined into the final change map.} 
		\label{fig:framework}
	\end{figure}

    As shown in Fig.~\ref{fig:framework}, Freq-RemoteVAR encodes
    the bi-temporal images into multi-scale features and fuses
    their appearance, difference, and correlation cues into
    conditional memory $\mathcal{M}$.
    Conditioned on this memory, the Frequency VAR Transformer
    generates mask tokens sequentially within frequency groups,
    following a low-to-high frequency order.
    SRCA aligns mask queries with multi-scale memory in a common
    image-space coordinate system, while CQC regulates evidence
    retrieval according to its estimated reliability.
    The generated tokens are mapped to codebook embeddings and
    decoded into the final change mask.

 \subsection{Preliminary: Visual Autoregressive Modeling}

Visual autoregressive modeling represents visual content as discrete tokens obtained by quantizing encoded features against a learned codebook. These tokens are mapped back to embeddings and decoded for reconstruction. Unlike next-token generation, which sequentially predicts flattened tokens, next-scale VAR~\cite{tian2024visual} generates token maps at progressively finer resolutions:
\begin{equation}
    p(r_1,\ldots,r_K\mid c)
    =
    \prod_{k=1}^{K}p(r_k\mid r_{<k},c),
    \label{eq:scale_ar}
\end{equation}
where $r_k$ denotes the token map at scale $k$, $r_{<k}$ contains preceding maps, and $c$ is the condition. This formulation organizes generation by spatial resolution, while regional structure and boundary variations may coexist within each scale. We instead organize change-mask prediction by frequency content.

\subsection{Frequency VAR: Next-Frequency Change Mask Prediction}
Freq-RemoteVAR uses the frequency composition of the target change mask to define the generation order. Lower-frequency components primarily capture broad change extent and connectivity, while higher-frequency components contribute finer contours and local variations. Unlike frequency-based feature enhancement, this formulation explicitly conditions finer-frequency prediction on preceding lower-frequency structure.

Let $\mathcal{T}=\{T_1,\ldots,T_K\}$ denote frequency-token groups ordered from low to high frequency. Our default setting uses $K=3$, corresponding to low-, mid-, and high-frequency groups. Given the bi-temporal condition tokens $C$, their joint distribution is factorized as
\begin{equation}
    p(\mathcal{T}\mid C)
    =
    \prod_{k=1}^{K}p(T_k\mid T_{<k},C),
    \label{eq:frequency_ar}
\end{equation}
where $T_{<k}$ contains all preceding frequency groups. Each group $T_k=\{t_{k,1},\ldots,t_{k,N_k}\}$ comprises $N_k$ discrete token indices, with within-group prediction defined as
\begin{equation}
    p(T_k\mid T_{<k},C)
    =
    \prod_{n=1}^{N_k}
    p(t_{k,n}\mid T_{<k},t_{k,<n},C),
    \label{eq:within_frequency_ar}
\end{equation}
where $t_{k,<n}$ denotes earlier tokens within group $k$. Thus, generation follows a low-to-high frequency order while maintaining within-group causality. Two-dimensional token coordinates are retained through positional encoding. The resulting hierarchy is determined by frequency content rather than token-map resolution, with bi-temporal image evidence guiding each prediction.

\subsection{Frequency-aware Mask Tokenization}

\begin{figure}[pos=!t,abovecap=6pt,belowcap=3pt]
		\centering
		\includegraphics[width=0.97\linewidth]{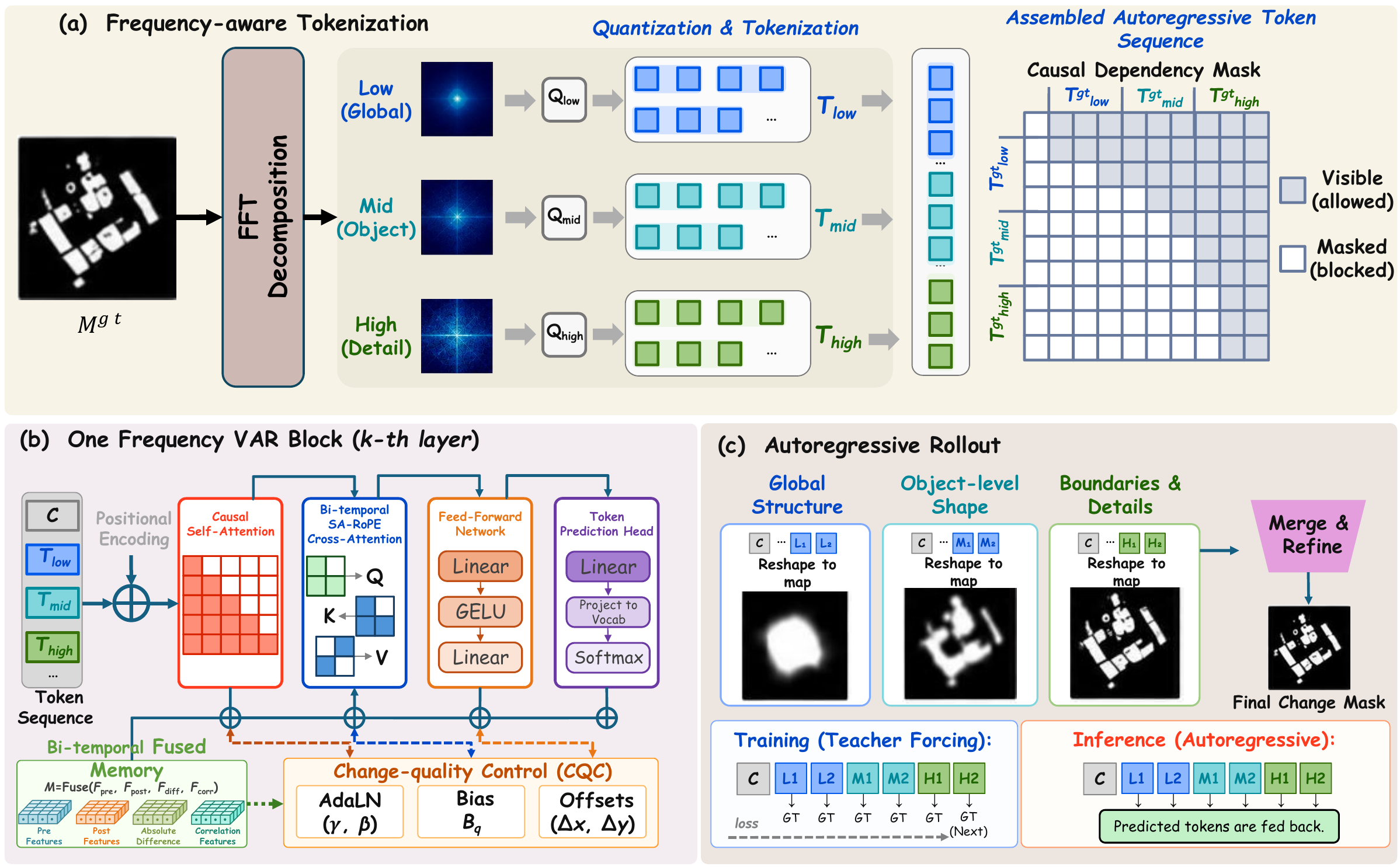}
		\caption{(a) Frequency-aware tokenization decomposes ground-truth change mask into low, mid, and high-frequency tokens and assembles them into a causal autoregressive sequence. 
			(b) A Frequency VAR block performs causal self-attention, Scale-Aligned RoPE Cross Attention, feed-forward transformation, and token prediction, with Change-quality Control providing adaptive modulation. 
			(c) The autoregressive rollout proceeds from lower- to higher-frequency token groups, progressively enriching coarse structural information with finer local variations before final mask reconstruction. } 
		\label{fig:methods_details}
	\end{figure}

Following discrete visual representation learning
\cite{van2017neural,esser2021taming}, we tokenize the target
change mask to construct supervision for autoregressive
prediction. As shown in Fig.~\ref{fig:methods_details}(a),
a lightweight mask encoder maps the ground-truth mask
$M^{gt}$ into a latent representation $f_m$, which is
decomposed into frequency bands:
\begin{equation}
    f_k
    =
    \mathcal{F}^{-1}
    \left(\mathcal{F}(f_m)\odot B_k\right),
    \qquad k\in\{low,mid,high\}.
    \label{eq:mask_frequency_decomposition}
\end{equation}
The selection masks $B_k$ are defined on the centered
Fourier plane, are non-overlapping, and jointly cover
the complete spectrum. The same frequency partition
is used for tokenization and frequency-consistency
supervision.

Each band-limited representation is converted into
discrete indices by a frequency-specific quantizer,
$T_k=Q_k(f_k)$. The resulting groups provide complementary
frequency-dependent representations, with lower frequencies
emphasizing coarse structure and higher frequencies
capturing finer spatial variations.
For reconstruction, the indices are mapped to codebook
embeddings and jointly decoded:
\begin{equation}
    \begin{aligned}
        E_k
        &=\mathrm{Lookup}(\mathcal{Z}_k,T_k),\\
        \tilde{M}
        &=D_m(E_{low},E_{mid},E_{high}),
    \end{aligned}
    \label{eq:tokenizer_reconstruction}
\end{equation}
where $\mathcal{Z}_k$ is the codebook for frequency band $k$
and $D_m$ is the mask decoder.

The tokenizer is optimized using reconstruction and
quantization objectives:
\begin{equation}
    \mathcal{L}_{tok}
    =
    \mathcal{L}_{rec}(\tilde{M},M^{gt})
    +
    \lambda_q\mathcal{L}_q,
    \label{eq:tokenizer_loss}
\end{equation}
where $\mathcal{L}_{rec}$ encourages accurate mask
reconstruction, $\mathcal{L}_q$ constrains the discrepancy
between continuous latent features and their quantized
embeddings, and $\lambda_q$ balances the two terms.

	\subsection{Bi-temporal Fusion Memory}
	
	To provide reliable image evidence for autoregressive change-mask generation, we construct a bi-temporal fused memory from the pre-change and post-change representations. 	As shown in Fig.~\ref{fig:framework}, directly using \(F_{pre}\) and \(F_{post}\) is insufficient to highlight change-sensitive cues. Therefore, we further construct two complementary bi-temporal representations: an absolute difference feature and a correlation feature. The absolute difference feature is computed as
	\begin{equation}
		F_{diff} = |F_{pre} - F_{post}|,
	\end{equation}
	which emphasizes local response discrepancies between the two temporal observations. Meanwhile, the correlation feature is defined as
	\begin{equation}
		F_{corr} = \mathrm{Corr}(F_{pre}, F_{post}),
	\end{equation}
	where \(\mathrm{Corr}(\cdot,\cdot)\) measures local feature correspondence between the pre- and post-change representations. Compared with simple differencing, the correlation feature helps capture temporal consistency and distinguish genuine changes from local misalignment or appearance variations.
	
	The bi-temporal fused memory is then obtained by integrating these four types of features:
	\begin{equation}
		\mathcal{M}
		=
		\mathrm{Fuse}
		\left(
		F_{pre},
		F_{post},
		F_{diff},
		F_{corr}
		\right),
	\end{equation}
	where $\mathrm{Fuse}(\cdot)$ denotes a lightweight fusion module. For notational simplicity, the scale index is omitted in Eqs.~(17)--(19); in implementation, the difference, correlation, and fusion operations are applied independently at each encoder scale, after which the projected multi-scale features are concatenated to construct $\mathcal{M}$. The memory $\mathcal{M}$ subsequently serves as the key and value source in SRCA, allowing the frequency tokens to continuously query change evidence from the bi-temporal image pair.

\subsection{Frequency VAR Transformer}

The Frequency VAR Transformer is designed to autoregressively
predict frequency-group tokens under bi-temporal image
conditions. Different from unconditional visual autoregressive
generation, our task requires the predicted mask tokens to be
grounded in the pre- and post-change images.

Therefore, we use the bi-temporal fused memory $\mathcal{M}$
as the condition source and project it into condition tokens
$C=\phi_c(\mathcal{M})$, where $\phi_c(\cdot)$ denotes a
lightweight projection layer.

During training, the frequency-token sequence is right-shifted
for teacher forcing. Each prediction position receives the
condition tokens $C$ and only the previously available
ground-truth frequency tokens, while the token to be predicted
and all future tokens remain invisible. During inference,
previously predicted tokens are recursively fed back into
the sequence. The causal dependency mask therefore guarantees
that each token is predicted exclusively from the bi-temporal
condition and preceding frequency tokens, avoiding target leakage.

As shown in Fig.~\ref{fig:methods_details}(b), each Frequency
VAR block consists of three main components: causal
self-attention, CQC-guided Scale-Aligned RoPE Cross Attention,
and a feed-forward network.
The token prediction heads are applied after stacking multiple
Frequency VAR blocks to project the final token representations
into the corresponding frequency token vocabularies.

Given the input token representation $X^{l}$ at the $l$-th
layer, causal self-attention first models dependencies among
frequency tokens:
\begin{equation}
    \hat{X}^{l}
    =
    X^{l}
    +
    \mathrm{CausalSA}
    \left(\mathrm{LN}(X^{l})\right).
    \label{eq:fvar_causal_sa}
\end{equation}
This operation allows lower-frequency tokens to provide
contextual guidance for subsequent higher-frequency token
prediction under the causal dependency mask.

For the $k$-th frequency group, CQC estimates memory
reliability and aggregates local bi-temporal evidence around
the query coordinates $p_{q,k}$. The resulting quality
descriptors are used to predict AdaLN modulation parameters
$\gamma_k^{l}$ and $\beta_k^{l}$, together with a raw spatial
offset $o_k^{l}$. The raw offset is converted into a bounded
displacement $\Delta p_k^{l}$, while the attention bias
$B_k^{l}$ is constructed from the estimated reliability
scores and coordinate distances rather than directly
predicted by CQC.

The AdaLN parameters modulate the normalized token features
$\hat{X}_k^{l}$ to obtain $\bar{X}_k^{l}$.
The detailed modulation, offset, and bias formulations are
provided in the next subsection, where the modulation input
$X_k$ corresponds to $\hat{X}_k^{l}$ here.

After that, the modulated tokens query the bi-temporal fused
memory through CQC-guided Scale-Aligned RoPE Cross Attention:
\begin{equation}
    \tilde{X}_k^{l}
    =
    \hat{X}_k^{l}
    +
    \mathrm{CrossAttn}_{\mathrm{CQC\text{-}SA\text{-}RoPE}}
    \left(
        \bar{X}_k^{l},
        \mathcal{M};
        B_k^{l},
        \Delta p_k^{l}
    \right).
    \label{eq:fvar_cross_attn}
\end{equation}

Here, $\mathcal{M}$ supplies bi-temporal image evidence,
SRCA aligns frequency queries with multi-scale memory in
a unified coordinate system, and CQC regulates feature
modulation, attention bias, and spatial offset to suppress
unreliable pseudo-change responses.

Finally, the output features are processed by a feed-forward
network:
\begin{equation}
    X_k^{l+1}
    =
    \tilde{X}_k^{l}
    +
    \mathrm{FFN}
    \left(\mathrm{LN}(\tilde{X}_k^{l})\right).
    \label{eq:fvar_ffn}
\end{equation}

After stacking $L$ Frequency VAR blocks, frequency-specific
prediction heads map the final representations to categorical
distributions over their codebooks. During inference, each
token is selected from its predicted distribution:
\begin{equation}
    \begin{aligned}
        P_k
        &=
        \mathrm{Softmax}
        \left(\mathrm{Head}_k(X_k^{L})\right),\\
        \hat{t}_{k,n}
        &=
        \arg\max_v P_k^{n,v},
    \end{aligned}
    \label{eq:fvar_prediction}
\end{equation}
where $k\in\{low,mid,high\}$ and $P_k^{n,v}$ denotes the
probability of codebook index $v$ at position $n$.
The distributions are evaluated sequentially during inference,
and $\hat{T}_k$ collects the generated indices within group $k$.

The predicted indices are subsequently converted to codebook
embeddings
$\hat{E}_k=\mathrm{Lookup}(\mathcal{Z}_k,\hat{T}_k)$
and decoded by $D_m$ to produce the final change mask.
    
\subsection{Scale-Aligned RoPE Cross-Attention with Change-Quality Control}

Conditioning frequency tokens on bi-temporal features requires
both spatial alignment and reliable evidence retrieval.
Frequency queries and multi-scale memory tokens originate
from grids with different resolutions, making their native
coordinates inconsistent. Moreover, spatially corresponding
features may still contain unreliable evidence caused by
seasonal variations, shadows, or slight misregistration.
SRCA addresses the coordinate inconsistency by mapping queries
and memory tokens into a common image-space coordinate system,
while CQC regulates feature modulation and local evidence
retrieval according to estimated reliability.

Given the $k$-th frequency-token representation
$X_k\in\mathbb{R}^{N_k\times d}$, we assign each query and
memory token an image-space coordinate:
\begin{equation}
    \begin{aligned}
        p_{q,k}^{i,j}
        &=
        \left(
            \frac{(i+0.5)H}{H_k},
            \frac{(j+0.5)W}{W_k}
        \right),\\
        p_{m,s}^{a,b}
        &=
        \left(
            \frac{(a+0.5)H}{H_s},
            \frac{(b+0.5)W}{W_s}
        \right),
    \end{aligned}
    \label{eq:srca_coordinates}
\end{equation}
where $H$ and $W$ denote the input image size,
$H_k$ and $W_k$ are the query-grid dimensions, and
$H_s$ and $W_s$ are the memory-grid dimensions at encoder
scale $s$. The coordinates from all memory scales are
collected in $p_m$, enabling positional interaction in
a shared spatial reference.

To assess the reliability of locally available evidence,
CQC predicts memory reliability scores and aggregates
quality-aware descriptors around the query coordinates:
\begin{equation}
    \begin{aligned}
        R
        &=\sigma(\psi_r(\mathcal{M})),\\
        z_k
        &=\mathrm{LocalAgg}(\mathcal{M},R,p_{q,k}),\\
        \{\gamma_k,\beta_k,o_k\}
        &=\psi_c(z_k),
    \end{aligned}
    \label{eq:cqc_quality}
\end{equation}
where $\psi_r$ and $\psi_c$ are lightweight prediction heads,
and $R\in[0,1]^{N_m}$ represents the estimated reliability
of memory tokens.
$\mathrm{LocalAgg}$ aggregates local evidence according to
the query coordinates and reliability scores.
The resulting descriptor produces AdaLN parameters
$\gamma_k,\beta_k\in\mathbb{R}^{N_k\times d}$ and a raw
spatial offset $o_k\in\mathbb{R}^{N_k\times2}$.

The offset adjusts the evidence-search position and is
bounded to avoid unstable displacement:
\begin{equation}
    \Delta p_k=\delta_k\tanh(o_k),
    \qquad
    \tilde{p}_{q,k}=p_{q,k}+\Delta p_k,
    \label{eq:cqc_offset}
\end{equation}
where $\delta_k$ bounds each offset component and sets
the spatial scale of the distance penalty below.
Meanwhile, the AdaLN parameters adapt the query features
to the estimated quality of the local evidence:
\begin{equation}
    \bar{X}_k
    =
    (1+\gamma_k)\odot\mathrm{LN}(X_k)+\beta_k.
    \label{eq:cqc_adaln}
\end{equation}
Here, the layer index is omitted, and $X_k$ corresponds
to the self-attention output $\hat{X}_k^{l}$ in the
Frequency VAR block.

We obtain the query, key, and value projections as
$Q_k=\bar{X}_kW_Q$,
$K=\mathcal{M}W_K$, and
$V=\mathcal{M}W_V$.
Scale-aligned rotary positional encoding is then applied
using the common image-space coordinates:
\begin{equation}
    \begin{aligned}
        Q_k'
        &=\mathrm{RoPE}(Q_k,\tilde{p}_{q,k}),\\
        K'
        &=\mathrm{RoPE}(K,p_m).
    \end{aligned}
    \label{eq:srca_rope}
\end{equation}
This positional interaction accounts for the spatial
correspondence between frequency queries and memory tokens
across different grid resolutions.

To further regulate evidence retrieval, we introduce a
quality-aware attention bias for query $q$ and memory token $m$:
\begin{equation}
    B_k^{q,m}
    =
    \lambda_r\log(R_m+\epsilon)
    -
    \lambda_d
    \frac{
        \left\|p_m-\tilde{p}_{q,k}^{q}\right\|_2^2
    }{\delta_k^2},
    \label{eq:cqc_attention_bias}
\end{equation}
where $\epsilon$ ensures numerical stability and
$\lambda_r,\lambda_d$ control the two bias terms.
The reliability term favors memory tokens with higher
estimated quality, while the distance penalty encourages
retrieval near the offset-adjusted query position.
The resulting CQC-guided SRCA output is
\begin{equation}
    \mathrm{Attn}_{\mathrm{CQC}}(Q_k,K,V)
    =
    \mathrm{Softmax}
    \left(
        \frac{Q_k'{K'}^{T}}{\sqrt{d_h}}+B_k
    \right)V,
    \label{eq:cqc_srca}
\end{equation}
where $d_h$ is the feature dimension of each attention head.
This output is incorporated through the residual update
in Eq.~\eqref{eq:fvar_cross_attn}.

SRCA and CQC thus provide complementary constraints on
bi-temporal evidence retrieval. SRCA establishes consistent
positional correspondence, while CQC adjusts the query
features, search position, and attention weights according
to local evidence quality. Together, they reduce the
influence of unreliable image differences on subsequent
frequency-token prediction.

\subsection{Training Objective}

The Frequency VAR Transformer is supervised by the discrete
frequency-token targets using
\begin{equation}
    \mathcal{L}_{AR}
    =
    \sum_{k\in\{low,mid,high\}}
    \frac{1}{N_k}
    \sum_{n=1}^{N_k}
    \mathrm{CE}(P_k^n,T_k^n),
    \label{eq:ar_loss}
\end{equation}
where $P_k^n$ and $T_k^n$ denote the predicted categorical
distribution and target codebook index, respectively.
To propagate mask-level supervision through token predictions,
we decode their expected codebook embeddings:
\begin{equation}
    \hat{M}
    =
    D_m\!\left(
        P_{low}\mathcal{Z}_{low},
        P_{mid}\mathcal{Z}_{mid},
        P_{high}\mathcal{Z}_{high}
    \right).
    \label{eq:soft_mask_reconstruction}
\end{equation}
During inference, expected embeddings are replaced by
hard token selection and codebook lookup.

We supervise $\hat{M}$ with binary cross-entropy
$\mathcal{L}_{bce}=\mathrm{BCE}(\hat{M},M^{gt})$
and Dice loss
$\mathcal{L}_{dice}
=1-\frac{2\sum_i\hat{M}_iM_i^{gt}+\epsilon}
{\sum_i\hat{M}_i+\sum_iM_i^{gt}+\epsilon}$,
which encourage pixel-wise accuracy and alleviate
foreground--background imbalance, respectively.
Here, $\epsilon$ ensures numerical stability.
We further enforce frequency consistency:
\begin{equation}
    \mathcal{L}_{freq}
    =
    \sum_{k\in\{low,mid,high\}}
    \left\|
        \mathcal{F}_k(\hat{M})
        -
        \mathcal{F}_k(M^{gt})
    \right\|_1,
    \label{eq:frequency_loss}
\end{equation}
where $\mathcal{F}_k$ extracts the corresponding frequency
component using the same selection mask $B_k$ as the tokenizer.
The overall objective is
\begin{equation}
    \begin{aligned}
        \mathcal{L}
        ={}&
        \lambda_{AR}\mathcal{L}_{AR}
        +\lambda_{bce}\mathcal{L}_{bce}\\
        &+\lambda_{dice}\mathcal{L}_{dice}
        +\lambda_{freq}\mathcal{L}_{freq},
    \end{aligned}
    \label{eq:total_loss}
\end{equation}
where the $\lambda$ terms balance the objectives.
The mask encoder, frequency-specific codebooks, Frequency VAR
Transformer, prediction heads, and mask decoder are jointly
optimized. The expected embeddings enable mask-level losses
to backpropagate through token-prediction distributions,
while $\mathcal{L}_{AR}$ directly supervises the discrete targets.

	\FloatBarrier
	\section{Experiments}
	\subsection{Datasets}
	
	Following the experimental protocol in~\cite{xie2026fsg}, we evaluate the proposed Freq-RemoteVAR on three widely used remote sensing change detection benchmarks, i.e., CDD, GZ-CD, and LEVIR-CD. 
	
	\textbf{CDD.} The CDD dataset~\cite{lebedev2018change} is collected from Google Earth and contains 11 pairs of raw remote sensing images with spatial resolutions ranging from 0.03 to 1 m/pixel. It involves both seasonal changes and disaster-induced changes, and is challenging due to shadows, illumination variations, cloud cover, and other non-semantic disturbances. Following the official pre-processed setting, we use 10,000 image pairs for training, 3,000 pairs for validation, and 3,000 pairs for testing. All image patches are resized or cropped to $256\times256$.
	
	\textbf{GZ-CD.} The GZ-CD dataset~\cite{peng2021semicdnet} is constructed from high-resolution Google Earth images collected through BIGEMAP, focusing on suburban areas of Guangzhou, China. The dataset contains season-varying bi-temporal image pairs with a spatial resolution of approximately 0.55 m/pixel, making it suitable for evaluating model robustness under seasonal appearance discrepancies. The original images are cropped into non-overlapping $256\times256$ patches. Following~\cite{xie2026fsg}, the final split contains 1,073 image pairs after filtering, including 751 pairs for training, 214 pairs for validation, and 108 pairs for testing.
	
	\textbf{LEVIR-CD.} The LEVIR-CD dataset~\cite{chen2020spatial} is a large-scale benchmark for building change detection. It consists of 637 pairs of very-high-resolution Google Earth images with a spatial resolution of 0.5 m/pixel. Each original image has a size of $1024\times1024$, and the temporal interval ranges from 5 to 14 years. The dataset mainly focuses on building changes across multiple cities in Texas, USA, and contains diverse urban expansion patterns. Following the official protocol, the original images are cropped into $256\times256$ patches, resulting in 7,120 training pairs, 1,024 validation pairs, and 2,048 testing pairs.

	\begin{table}[pos=!htbp]
\centering
\caption{Quantitative comparison on three change detection datasets.}
\label{tab:sota_comparison}
\setlength{\tabcolsep}{1.5pt}
\renewcommand{\arraystretch}{1.00}
\begin{tabular*}{\linewidth}{@{\extracolsep{\fill}}llcccccc}
\toprule
Methods & Ref. & Dataset & Pre. & Rec. & F1 & IoU & OA \\
\midrule
\multirow{3}{*}{BIT~\cite{chen2022remote}} & \multirow{3}{*}{TGRS'21} & CDD & 90.67 & 86.44 & 88.50 & 79.37 & 97.26 \\
 &  & GZ-CD & 82.90 & 80.68 & 81.77 & 69.16 & 90.63 \\
 &  & LEVIR-CD & 89.24 & 89.37 & 89.30 & 80.67 & 98.88 \\
\addlinespace[1pt]
\multirow{3}{*}{ChangeFormer~\cite{bandara2022transformer}} & \multirow{3}{*}{IGARSS'22} & CDD & 91.54 & 89.31 & 90.41 & 82.50 & 97.68 \\
 &  & GZ-CD & 84.59 & 84.28 & 84.43 & 73.06 & 91.90 \\
 &  & LEVIR-CD & 91.97 & 89.03 & 90.48 & 82.62 & 99.02 \\
\addlinespace[1pt]
\multirow{3}{*}{DMINet~\cite{feng2023change}} & \multirow{3}{*}{TGRS'23} & CDD & 92.45 & 90.72 & 91.58 & 84.47 & 97.96 \\
 &  & GZ-CD & 89.04 & 86.52 & 87.76 & 78.19 & 93.71 \\
 &  & LEVIR-CD & 92.27 & 88.94 & 90.57 & 82.77 & 99.03 \\
\addlinespace[1pt]
\multirow{3}{*}{AMTNet~\cite{liu2023attention}} & \multirow{3}{*}{ISPRS'23} & CDD & 92.32 & 92.91 & 92.61 & 86.24 & 98.19 \\
 &  & GZ-CD & 88.51 & 85.27 & 86.86 & 76.77 & 93.28 \\
 &  & LEVIR-CD & 91.34 & 89.76 & 90.54 & 82.72 & 99.02 \\
\addlinespace[1pt]
\multirow{3}{*}{WS-Net++~\cite{xiong2024wavelet}} & \multirow{3}{*}{TGRS'24} & CDD & 92.95 & 93.21 & 93.08 & 87.06 & 98.31 \\
 &  & GZ-CD & 88.32 & 85.76 & 87.02 & 77.02 & 93.33 \\
 &  & LEVIR-CD & 92.11 & \secondbest{90.05} & 91.07 & 83.60 & 99.08 \\
\addlinespace[1pt]
\multirow{3}{*}{CDNeXt~\cite{wei2024robust}} & \multirow{3}{*}{JAG'24} & CDD & 92.76 & 93.87 & 93.31 & 87.46 & 98.35 \\
 &  & GZ-CD & 89.21 & 86.34 & 87.75 & 78.17 & 93.72 \\
 &  & LEVIR-CD & 92.15 & 89.69 & 90.91 & 83.33 & 99.06 \\
\addlinespace[1pt]
\multirow{3}{*}{FTransDF-Net~\cite{li2025dual}} & \multirow{3}{*}{JAG'25} & CDD & 93.10 & 94.03 & 93.56 & 87.90 & 98.42 \\
 &  & GZ-CD & 89.13 & 87.45 & 88.28 & 79.02 & 93.95 \\
 &  & LEVIR-CD & 92.19 & 89.73 & 90.94 & 83.39 & 99.07 \\
\addlinespace[1pt]
\multirow{3}{*}{ConvFormer~\cite{yang2025convformer}} & \multirow{3}{*}{TGRS'25} & CDD & 93.35 & 93.19 & 93.28 & 87.41 & 98.36 \\
 &  & GZ-CD & 89.11 & 87.58 & 88.33 & 79.11 & 93.97 \\
 &  & LEVIR-CD & 92.14 & 89.69 & 90.90 & 83.32 & 99.06 \\
\addlinespace[1pt]
\multirow{3}{*}{FSG-Net~\cite{xie2026fsg}} & \multirow{3}{*}{TGRS'26} & CDD & 93.33 & \secondbest{95.01} & 94.16 & 88.96 & 98.56 \\
 &  & GZ-CD & \secondbest{90.85} & \secondbest{88.20} & \secondbest{89.51} & \secondbest{81.01} & \secondbest{94.61} \\
 &  & LEVIR-CD & \secondbest{92.53} & 90.04 & \secondbest{91.27} & \secondbest{83.94} & \secondbest{99.10} \\
\addlinespace[1pt]
\multirow{3}{*}{CDMask~\cite{ma2026cdmask}} & \multirow{3}{*}{TGRS'26} & CDD & \textbf{96.55} & 94.92 & \secondbest{95.73} & \secondbest{91.81} & \secondbest{98.95} \\
 &  & GZ-CD & 83.95 & 85.44 & 84.69 & 73.44 & 92.70 \\
 &  & LEVIR-CD & 89.42 & 89.45 & 89.44 & 80.89 & 98.92 \\
\midrule
\multirow{3}{*}{\textbf{Freq-RemoteVAR}} & \multirow{3}{*}{Ours} & CDD & \secondbest{95.95} & \textbf{96.13} & \textbf{96.04} & \textbf{92.38} & \textbf{99.07} \\
 &  & GZ-CD & \textbf{91.22} & \textbf{89.54} & \textbf{90.37} & \textbf{82.44} & \textbf{94.97} \\
 &  & LEVIR-CD & \textbf{92.78} & \textbf{90.98} & \textbf{91.87} & \textbf{84.96} & \textbf{99.17} \\
\bottomrule
\end{tabular*}
\par\vspace{3pt}
\noindent\begin{minipage}{\linewidth}
\PRcaptionfont
The best results are highlighted in \textbf{bold}, and the second-best results are underlined.
\end{minipage}
\end{table}

\subsection{Experimental Setup}
	
\textbf{Implementation details.}
All experiments are implemented in PyTorch. For a fair comparison, we follow the experimental protocol of~\cite{xie2026fsg} as closely as possible. All input image pairs and ground-truth masks are resized or cropped to $256\times256$. The proposed Freq-RemoteVAR is optimized using the AdamW optimizer with a weight decay of 0.01. The initial learning rate is set to $1\times10^{-4}$ for the pre-trained backbone and $1\times10^{-3}$ for the newly introduced tokenizer, Frequency VAR Transformer, decoder, and prediction heads. The learning rate is gradually decayed to $1\times10^{-6}$ using a cosine annealing schedule. The model is trained for 100 epochs on each dataset.

\textbf{Evaluation metrics.}
Following common practice in remote sensing change detection, we adopt five widely used evaluation metrics, including Precision (Pre.), Recall (Rec.), F1-score (F1), Intersection over Union (IoU), and Overall Accuracy (OA). 
    

	\subsection{Comparison With State-of-the-art Methods}
	
	We compare the proposed Freq-RemoteVAR with representative remote sensing change detection methods, including BIT~\cite{chen2022remote}, ChangeFormer~\cite{bandara2022transformer}, DMINet~\cite{feng2023change}, AMTNet~\cite{liu2023attention}, WS-Net++~\cite{xiong2024wavelet}, CDNeXt~\cite{wei2024robust}, FTransDF-Net~\cite{li2025dual}, ConvFormer~\cite{yang2025convformer}, FSG-Net~\cite{xie2026fsg}, and CDMask~\cite{ma2026cdmask}. These methods cover different technical paradigms, including CNN-Transformer hybrid modeling, bi-temporal feature interaction, frequency-domain enhancement, temporospatial attention, gated fusion, and mask-query-based change prediction. 
	
\textbf{Quantitative Comparison.} 
Table~\ref{tab:sota_comparison} reports the quantitative
results on CDD, GZ-CD, and LEVIR-CD. Freq-RemoteVAR achieves
the best F1-score, IoU, and OA on all three datasets,
together with the highest Recall on CDD and the highest
Precision and Recall on GZ-CD and LEVIR-CD. These results
demonstrate consistent improvements over both early
Transformer-based methods and recent competitive approaches.

On CDD, Freq-RemoteVAR achieves 96.04\% F1-score and
92.38\% IoU, exceeding CDMask by 0.31 and 0.57 percentage
points, respectively. Although CDMask obtains the highest
Precision of 96.55\%, its Recall of 94.92\% is lower than
the 96.13\% achieved by our method. Compared with FSG-Net,
Freq-RemoteVAR also improves F1-score and IoU by 1.88 and
3.42 percentage points. This combination of higher Recall
and the best F1-score indicates more complete change
detection with a favorable balance between false alarms
and missed changes.

On GZ-CD, seasonal appearance variations and complex
backgrounds make change detection challenging.
Freq-RemoteVAR achieves the best results across all five
metrics, including 90.37\% F1-score and 82.44\% IoU.
Compared with the strongest competing method, FSG-Net,
these represent gains of 0.86 and 1.43 percentage points,
respectively. The improvements support the effectiveness
of the proposed framework under the challenging
bi-temporal conditions represented in this dataset.

On LEVIR-CD, competing methods exhibit relatively close
performance on building change detection. Nevertheless,
Freq-RemoteVAR achieves 91.87\% F1-score and 84.96\% IoU,
surpassing the second-best FSG-Net by 0.60 and 1.02
percentage points, respectively. It also increases Recall
from the 90.05\% achieved by WS-Net++ to 90.98\%, while
obtaining the highest Precision of 92.78\%. These results
indicate improved completeness of detected building
changes while maintaining high precision.

	\begin{figure}[pos=!htbp,abovecap=6pt,belowcap=3pt]
		\centering
		\includegraphics[width=\textwidth]{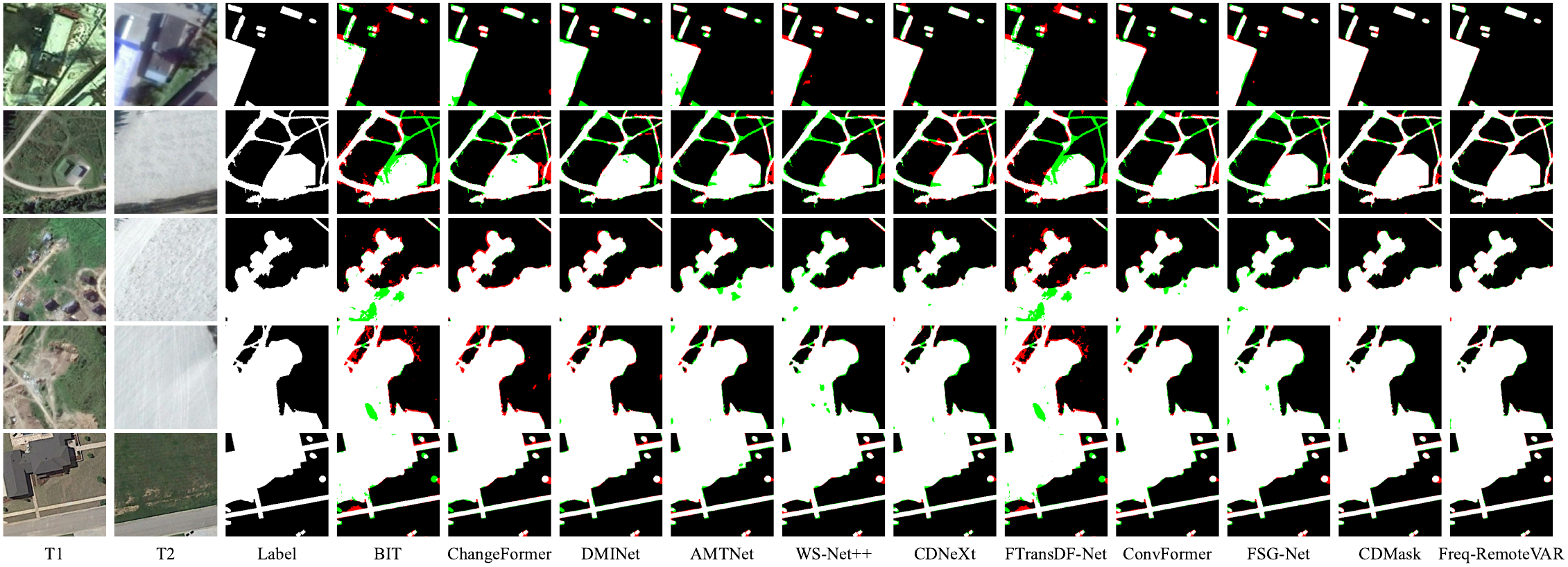}
		\caption{Qualitative comparison on the CDD dataset. White, black, red, and green pixels denote true positives, true negatives, false positives, and false negatives, respectively.}
		\label{fig:qual_cdd}
	\end{figure}

	\begin{figure}[pos=!htbp,abovecap=6pt,belowcap=3pt]
		\centering
		\includegraphics[width=\textwidth]{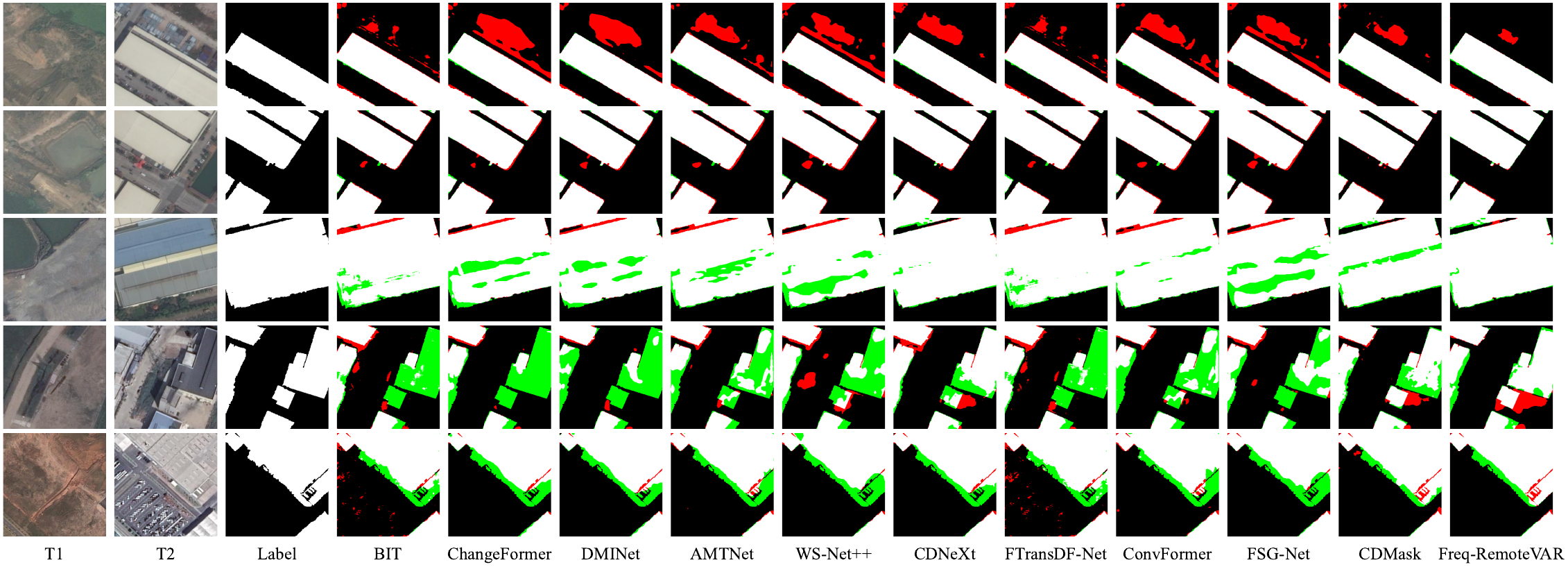}
		\caption{Qualitative comparison on the GZ-CD dataset. White, black, red, and green pixels denote true positives, true negatives, false positives, and false negatives, respectively.}
		\label{fig:qual_gzcd}
	\end{figure}

    \begin{figure}[pos=!htbp,abovecap=6pt,belowcap=3pt]
		\centering
		\includegraphics[width=\textwidth]{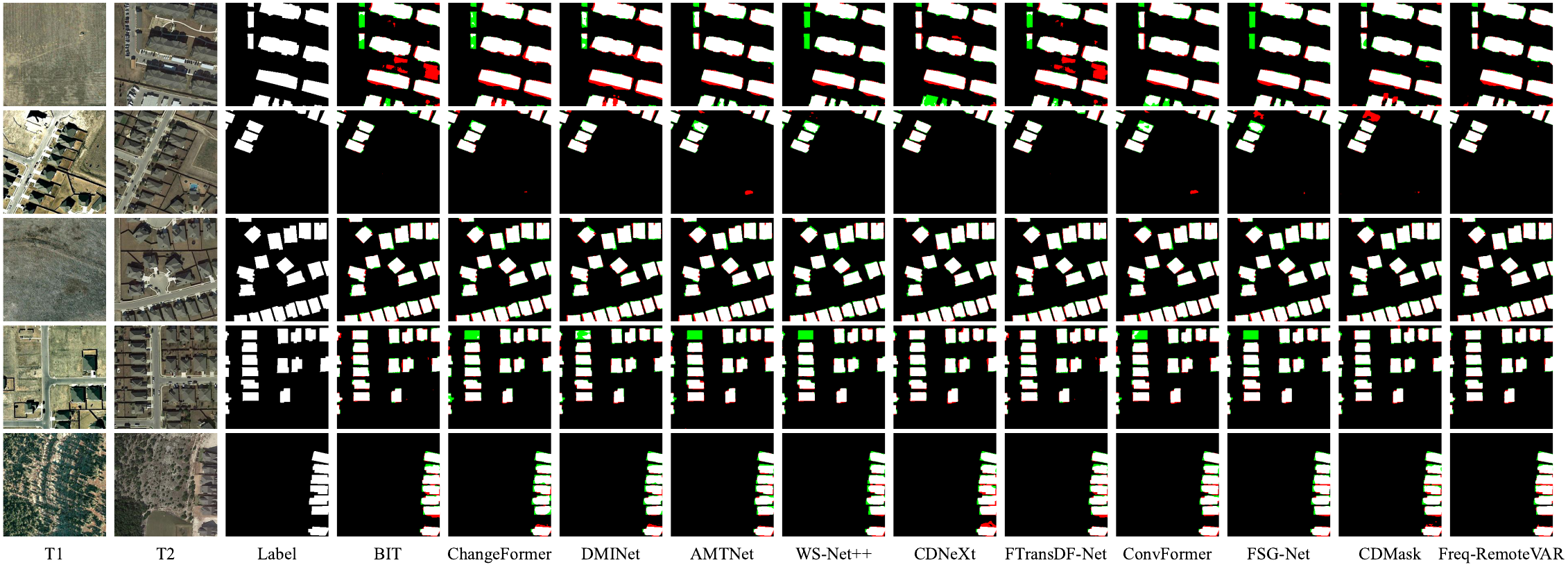}
		\caption{Qualitative comparison on the LEVIR-CD dataset. White, black, red, and green pixels denote true positives, true negatives, false positives, and false negatives, respectively.}
		\label{fig:qual_levir}
	\end{figure}

	\textbf{Qualitative Comparison.} To further evaluate the visual quality of different methods, we provide qualitative comparisons on the CDD, GZ-CD, and LEVIR-CD datasets in Figs.~\ref{fig:qual_cdd}--\ref{fig:qual_levir}. Following the common visualization protocol in change detection, white pixels denote true positives, black pixels denote true negatives, red pixels denote false positives, and green pixels denote false negatives. Therefore, red regions indicate pseudo-change responses, while green regions correspond to missed changed areas.
	
	As shown in Fig.~\ref{fig:qual_cdd}, CDD contains complex background disturbances, including illumination variations, shadows, vegetation changes, and irregular land-cover appearances. BIT and ChangeFormer tend to produce obvious red false alarms in unchanged areas, while several later methods reduce part of these false positives but still suffer from incomplete boundaries or local missing regions. Although FSG-Net and CDMask generate cleaner predictions than earlier baselines, they still show visible red or green errors in some irregular regions. In contrast, Freq-RemoteVAR produces more compact and continuous change masks, with fewer isolated false positives and better preserved changed structures. This visual comparison is consistent with the quantitative advantage of our method in Recall, F1-score, and IoU on CDD.
	
	Fig.~\ref{fig:qual_gzcd} presents visual comparisons on GZ-CD. Due to strong seasonal differences and complex suburban backgrounds, several methods either over-detect background variations as changes or miss elongated and irregular changed regions. This phenomenon is especially visible in the green missing regions and red pseudo-change responses of some competing methods. Compared with these methods, Freq-RemoteVAR better suppresses pseudo changes in unchanged backgrounds while maintaining the integrity of changed objects.
	
	For LEVIR-CD shown in Fig.~\ref{fig:qual_levir}, the main challenge lies in accurately detecting building changes with dense layouts, sharp boundaries, and diverse object scales. Some competing methods produce fragmented building masks, inaccurate contours, or small false alarms around unchanged buildings. FSG-Net and CDMask provide competitive results, but their predictions still contain visible boundary errors and local missing regions in several samples. Benefiting from frequency-aware progressive generation, Scale-Aligned RoPE Cross Attention, and Change-quality Control, Freq-RemoteVAR produces cleaner building masks with more accurate boundary delineation.

\FloatBarrier
\section{Ablation Study}

\begin{figure}[pos=!htbp,abovecap=6pt,belowcap=3pt]
		\centering
		\includegraphics[width=\linewidth]{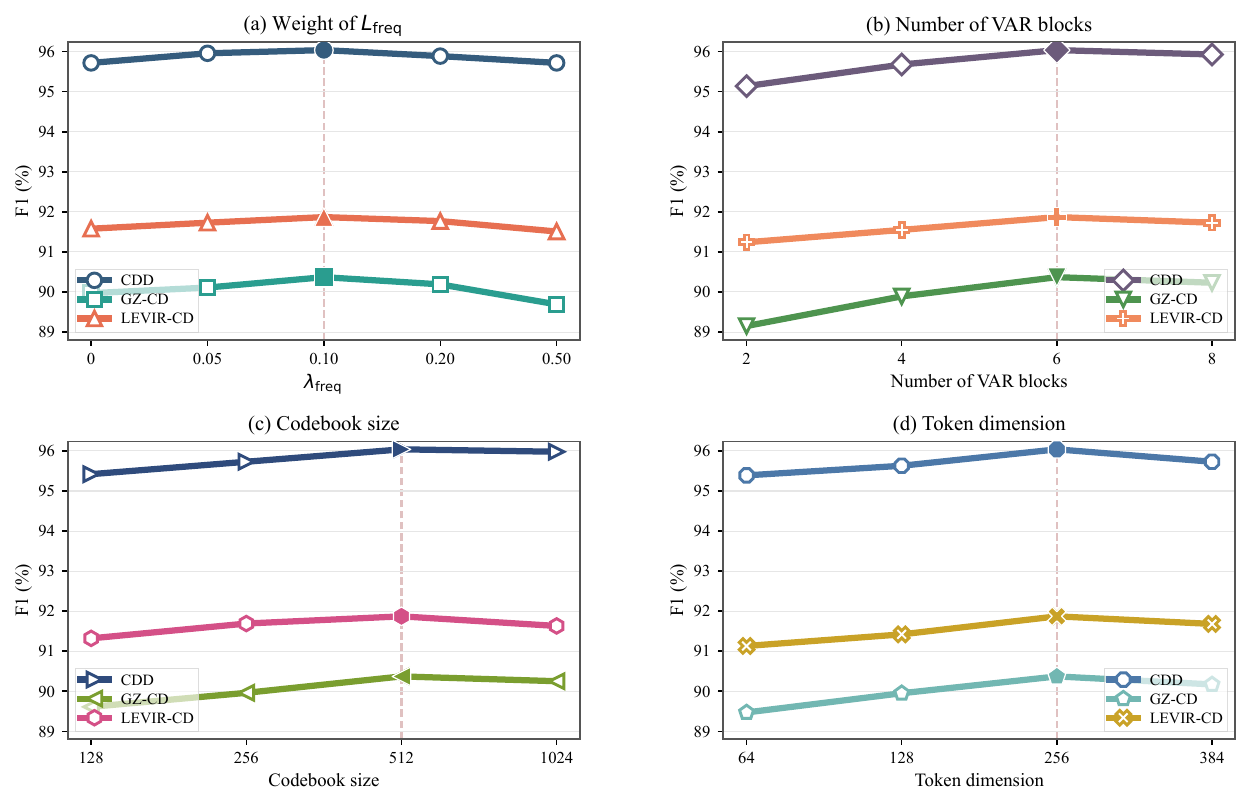}
		\caption{Parameter sensitivity analysis of Freq-RemoteVAR under different hyper-parameter settings on CDD, GZ-CD, and LEVIR-CD. The investigated parameters include the weight of frequency consistency loss, the number of VAR blocks, the codebook size, and the token dimension.}
		\label{fig:para_vis}
	\end{figure}

	\begin{figure}[pos=!htbp,abovecap=6pt,belowcap=3pt]
		\centering
		\includegraphics[width=\linewidth]{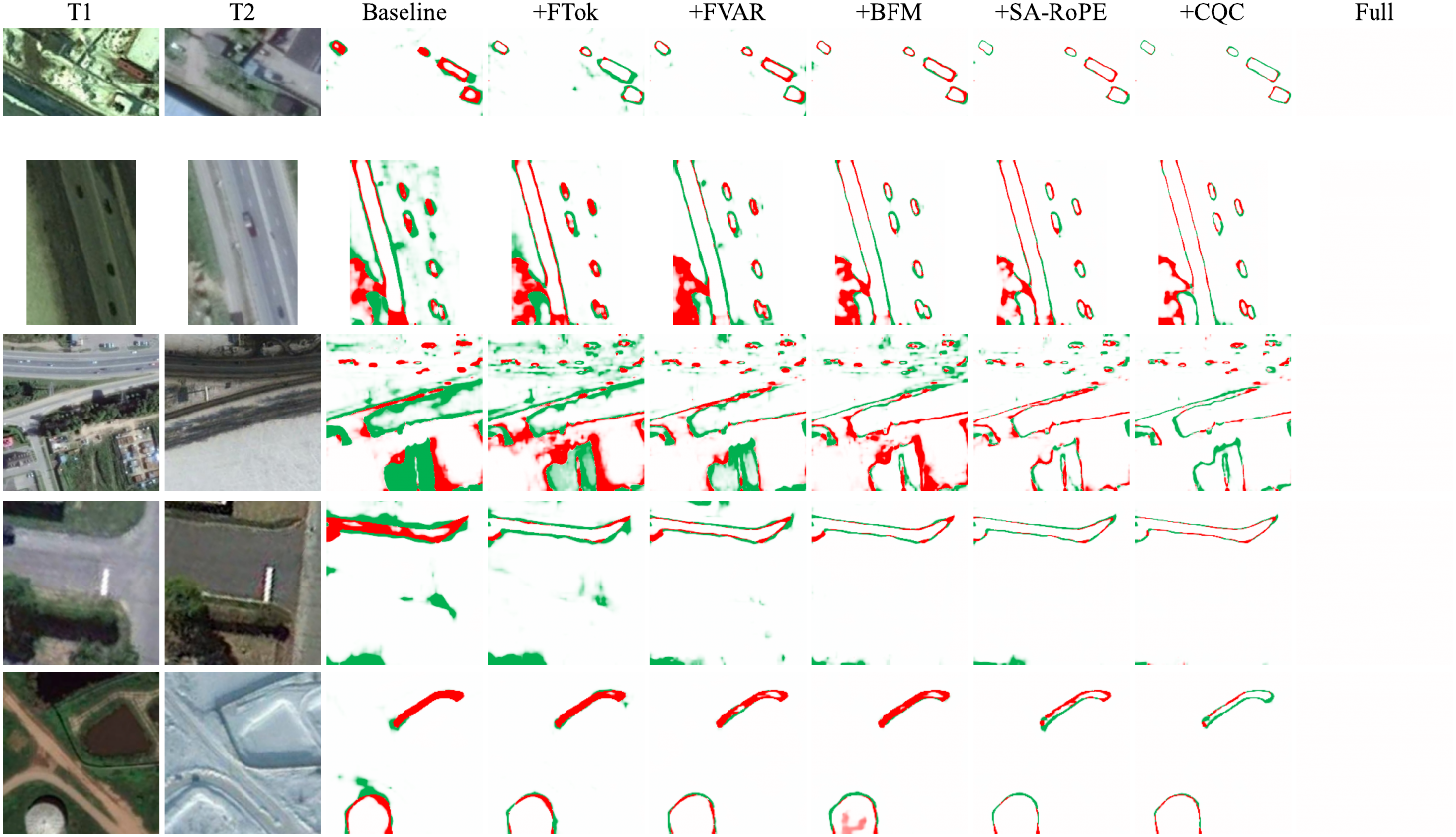}
		\caption{Module-wise response deviation visualization. Red and green regions indicate stronger and weaker change responses than the full model, respectively. Since the full model is used as the reference, its deviation map is nearly blank.}
		\label{fig:abs_vis}
	\end{figure}
	
\subsection{Module-wise Ablation}

Table~\ref{tab:module_ablation} presents cumulative module-wise
ablations on CDD, GZ-CD, and LEVIR-CD. The baseline uses a
Siamese encoder with simple bi-temporal feature fusion and
directly predicts the change mask. The full model improves
F1-score over this baseline by 2.62, 3.52, and 1.55 percentage
points on the three datasets, respectively.
In the FTok-only setting, lightweight heads independently
predict the three frequency-token groups, which are
reconstructed using the same frequency-token mask decoder
without cross-frequency autoregressive dependency.

\begin{table}[pos=!htbp]
		\centering
		\caption{Module-wise ablation study of Freq-RemoteVAR.}
		\label{tab:module_ablation}
		\setlength{\tabcolsep}{0.6pt}
		\renewcommand{\arraystretch}{1.05}
		\begin{tabular*}{\linewidth}{@{\extracolsep{\fill}}lcccccccccccc@{}}
			\toprule
			\multirow{2}{*}{Settings}
			& \multirow{2}{*}{FTok}
			& \multirow{2}{*}{FVAR}
			& \multirow{2}{*}{BFM}
			& \multirow{2}{*}{SRCA}
			& \multirow{2}{*}{CQC}
			& \multirow{2}{*}{$L_{\mathrm{freq}}$}
			& \multicolumn{2}{c}{CDD}
			& \multicolumn{2}{c}{GZ-CD}
			& \multicolumn{2}{c}{LEVIR-CD} \\
			\cmidrule(lr){8-9}\cmidrule(lr){10-11}\cmidrule(lr){12-13}
			& & & & & &
			& F1 & IoU
			& F1 & IoU
			& F1 & IoU \\
			\midrule

			Baseline
			& -- & -- & -- & -- & -- & --
			& 93.42 & 87.65
			& 86.85 & 76.76
			& 90.32 & 82.35 \\

			+ FTok
			& Y & -- & -- & -- & -- & --
			& 94.16 & 88.96
			& 87.64 & 78.00
			& 90.79 & 83.13 \\

			+ FVAR
			& Y & Y & -- & -- & -- & --
			& 94.53 & 89.63
			& 88.37 & 79.16
			& 91.14 & 83.72 \\

			+ BFM
			& Y & Y & Y & -- & -- & --
			& 95.21 & 90.86
			& 88.86 & 79.95
			& 91.42 & 84.20 \\

			+ SRCA
			& Y & Y & Y & Y & -- & --
			& 95.52 & 91.42
			& 89.41 & 80.85
			& 91.56 & 84.43 \\

			+ CQC
			& Y & Y & Y & Y & Y & --
			& 95.81 & 91.96
			& 89.87 & 81.60
			& 91.69 & 84.66 \\

			Full
			& Y & Y & Y & Y & Y & Y
			& \textbf{96.04} & \textbf{92.38}
			& \textbf{90.37} & \textbf{82.44}
			& \textbf{91.87} & \textbf{84.96} \\

			\bottomrule
		\end{tabular*}
	\end{table}

Introducing FTok improves F1-score by 0.74, 0.79, and 0.47
percentage points, supporting the use of frequency-decomposed
token targets. Adding the Frequency VAR Transformer yields
further gains across all datasets, consistent with the
benefit of modeling dependencies among frequency tokens.
The bi-temporal fused memory further improves performance,
suggesting that the additional image evidence supports
frequency-token generation.

SRCA increases F1-score by 0.31, 0.55, and 0.14 percentage
points over the preceding configuration, while CQC provides
further gains of 0.29, 0.46, and 0.13 percentage points.
These results support the contribution of spatially aligned
evidence retrieval and quality-aware modulation.
Both modules show larger gains on GZ-CD, which contains
challenging seasonal variations and background disturbances.
Finally, frequency-consistency supervision produces the
best results on all three datasets, demonstrating its
additional benefit within the complete framework.

Fig.~\ref{fig:abs_vis} visualizes the pixel-wise differences
between each ablated model's predicted change probabilities
and those of the full model. Red and green indicate stronger
and weaker responses relative to the full model, respectively.
As components are introduced, the displayed deviations
generally become more spatially concentrated, illustrating
where the predictions approach those of the complete
framework.

\subsection{Parameter Sensitivity Analysis}
	
To evaluate the robustness of Freq-RemoteVAR under different hyper-parameter settings, we conduct parameter sensitivity analysis on CDD, GZ-CD, and LEVIR-CD. As shown in Fig.~\ref{fig:para_vis}, we investigate four important hyper-parameters, including the weight of the frequency consistency loss \(\lambda_{\mathrm{freq}}\), the number of VAR blocks, the codebook size, and the token dimension. Overall, Freq-RemoteVAR shows stable performance within a reasonable parameter range, and the default configuration consistently achieves the best or near-best F1-score on the three datasets.

For the frequency consistency loss, the best performance is obtained when \(\lambda_{\mathrm{freq}}=0.10\). A smaller weight weakens the frequency-domain regularization, while an overly large weight may over-constrain the prediction and interfere with the main change detection objective. For the number of VAR blocks, the performance improves as the depth increases from 2 to 6, but slightly decreases when using 8 blocks, indicating that excessive depth may introduce redundancy and optimization difficulty. For the codebook size, 512 provides the best balance between token representation capacity and quantization stability. For the token dimension, 256 achieves the best overall performance, suggesting that a moderate embedding dimension is sufficient for frequency autoregressive change modeling.

\subsection{Autoregressive Strategy Ablation}
Table~\ref{tab:controlled_ar_comparison} compares one-shot prediction, parallel frequency prediction, RemoteVAR, a controlled spatial-scale AR variant, and Freq-RemoteVAR. The controlled spatial-scale AR variant retains the main architecture and bi-temporal modeling components but organizes prediction by spatial scale. These settings separate the effects of prediction organization from those of the supporting components.

\begin{table}[pos=htbp]
        \centering
        \caption{Controlled comparison of different prediction paradigms.}
        \label{tab:controlled_ar_comparison}
        \setlength{\tabcolsep}{1.0pt}
        \renewcommand{\arraystretch}{1.00}
        \renewcommand{\tabularxcolumn}[1]{m{#1}}
        \begin{tabularx}{\linewidth}{@{}
            >{\hsize=.9\hsize\linewidth=\hsize\raggedright\arraybackslash}X
            >{\hsize=1.1\hsize\linewidth=\hsize\raggedright\arraybackslash}X
            cc cc cc@{}}
            \toprule
            \multirow{2}{*}{Method}
            & \multirow{2}{*}{\shortstack[l]{Target / Prediction\\Organization}}
            & \multicolumn{2}{c}{CDD}
            & \multicolumn{2}{c}{GZ-CD}
            & \multicolumn{2}{c}{LEVIR-CD} \\
            \cmidrule(lr){3-4}\cmidrule(lr){5-6}\cmidrule(lr){7-8}
            & & F1 & IoU & F1 & IoU & F1 & IoU \\
            \midrule
    
            One-shot Prediction
            & Direct dense prediction
            & 94.65 & 89.84
            & 88.53 & 79.42
            & 91.01 & 83.50 \\
    
            Parallel Frequency
            & Frequency groups without AR
            & 95.33 & 91.08
            & 89.16 & 80.44
            & 91.35 & 84.08 \\
    
            RemoteVAR
            & Spatial-scale autoregression
            & 95.32 & 91.06
            & 89.24 & 80.57
            & 91.33 & 84.04 \\
    
            Spatial-scale AR (Ours)
            & Controlled spatial-scale AR
            & 95.53 & 91.44
            & 89.42 & 80.86
            & 91.42 & 84.20 \\
    
            \midrule
    
            \textbf{Freq-RemoteVAR}
            & \textbf{Frequency-ordered autoregression}
            & \textbf{96.04} & \textbf{92.38}
            & \textbf{90.37} & \textbf{82.44}
            & \textbf{91.87} & \textbf{84.96} \\
    
            \bottomrule
        \end{tabularx}
    \end{table}

Parallel frequency prediction improves upon one-shot prediction on all three datasets, supporting the use of frequency-decomposed targets. Introducing spatial-scale autoregression yields further gains, while frequency-ordered autoregression achieves the best results. Compared with the controlled spatial-scale AR variant, Freq-RemoteVAR improves F1 by 0.51, 0.95, and 0.45 percentage points on CDD, GZ-CD, and LEVIR-CD, respectively. Relative to RemoteVAR, the corresponding F1 gains are 0.72, 1.13, and 0.54 percentage points, with IoU gains of 1.32, 1.87, and 0.92 percentage points. These comparisons support the benefit of conditioning finer-frequency prediction on previously generated lower-frequency structure beyond spatial-scale autoregression alone.

\begin{figure}[pos=!t,abovecap=6pt,belowcap=3pt]
		\centering
		\includegraphics[width=\linewidth]{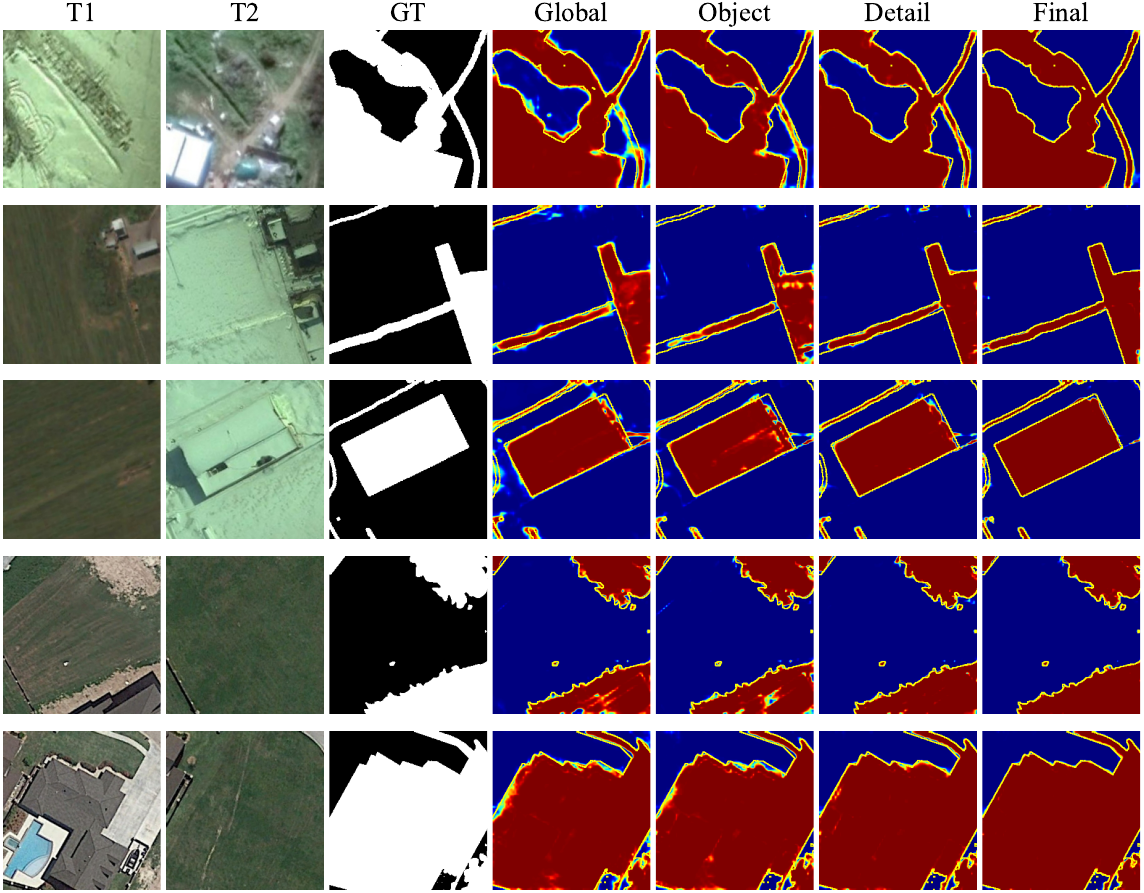}
		\caption{Visualization of progressive frequency-aware change responses. \textit{Global}, \textit{Object}, and \textit{Detail} correspond to low-, mid-, and high-frequency responses, respectively. Their fusion produces the final change prediction.}
		\label{fig:varfreq_vis}
	\end{figure}

To further visualize the progressive generation behavior of Freq-RemoteVAR, we present the frequency-aware change response maps in Fig.~\ref{fig:varfreq_vis}. The global response provides a coarse but stable change layout, the object-level response enhances the integrity of changed regions, and the detail response further refines boundaries and local structures. The final prediction integrates these complementary responses and produces a more complete change mask. This visualization intuitively supports the effectiveness of the low-to-high frequency autoregressive generation strategy.

\subsection{Causal Mask Analysis}
	
We further analyze the effect of different causal mask designs in Table~\ref{tab:causal_mask}. Without a causal mask, the model cannot explicitly constrain the generation dependency among frequency components, resulting in inferior performance. The independent frequency setting improves the results by separating different frequency groups, but it still lacks cross-frequency causal dependency. The spatial causal mask further improves the performance, suggesting that coarse-to-fine autoregressive constraints are beneficial for change detection.

\begin{table}[pos=htbp]
		\centering
		\caption{Causal mask analysis of frequency autoregressive modeling.}
		\label{tab:causal_mask}
		\setlength{\tabcolsep}{3.0pt}
		\renewcommand{\arraystretch}{1.00}
		\begin{tabular}{@{}lcccccc@{}}
			\toprule
			Mask
			& \multicolumn{2}{c}{CDD}
			& \multicolumn{2}{c}{GZ-CD}
			& \multicolumn{2}{c}{LEVIR-CD} \\
			\cmidrule(lr){2-3}\cmidrule(lr){4-5}\cmidrule(lr){6-7}
			& F1 & IoU
			& F1 & IoU
			& F1 & IoU \\
			\midrule
			
			No mask
			& 95.28 & 90.99
			& 89.05 & 80.26
			& 91.26 & 83.92 \\
			
			Indep. freq.
			& 95.49 & 91.37
			& 89.31 & 80.68
			& 91.41 & 84.18 \\
			
			Spatial causal
			& 95.67 & 91.70
			& 89.68 & 81.29
			& 91.55 & 84.42 \\
			
			Freq. causal
			& \textbf{96.04} & \textbf{92.38}
			& \textbf{90.37} & \textbf{82.44}
			& \textbf{91.87} & \textbf{84.96} \\
			
			\bottomrule
		\end{tabular}
		
	\end{table}

Compared with these alternatives, the proposed frequency causal mask achieves the best results, with F1-scores of 96.04\%, 90.37\%, and 91.87\% on CDD, GZ-CD, and LEVIR-CD, respectively. Compared with the spatial causal mask, it improves the F1-score by 0.37\%, 0.69\%, and 0.32\% on the three datasets. This verifies that the key advantage of Freq-RemoteVAR does not simply come from using autoregressive modeling, but from constructing a frequency-aware causal dependency that matches the intrinsic generation process of change masks. By enforcing low-frequency structures to guide mid- and high-frequency prediction, the frequency causal mask helps the model produce more complete change regions and more accurate boundaries.

\subsection{Frequency Partition Granularity Analysis}

We further investigate the effect of frequency partition granularity by varying the number of frequency groups from one to four while keeping the remaining architecture and training settings unchanged. As shown in Table~\ref{tab:freq_partition}, performance consistently improves from one to three bands on all three datasets.

\begin{table}[pos=htbp]
        \centering
        \caption{Effect of frequency partition granularity on CDD, GZ-CD, and LEVIR-CD.}
        \label{tab:freq_partition}
        \renewcommand{\arraystretch}{1.00}
        \setlength{\tabcolsep}{3.0pt}
        \begin{tabular}{@{}lcccccc@{}}
            \toprule
            Frequency Groups
            & \multicolumn{2}{c}{CDD}
            & \multicolumn{2}{c}{GZ-CD}
            & \multicolumn{2}{c}{LEVIR-CD} \\
            \cmidrule(lr){2-3}\cmidrule(lr){4-5}\cmidrule(lr){6-7}
            & F1 & IoU & F1 & IoU & F1 & IoU \\
            \midrule
    
            1 band
            & 95.04 & 90.55
            & 88.62 & 79.57
            & 91.02 & 83.52 \\
    
            2 bands
            & 95.61 & 91.59
            & 89.59 & 81.14
            & 91.46 & 84.26 \\
    
            \textbf{3 bands}
            & \textbf{96.04} & \textbf{92.38}
            & \textbf{90.37} & \textbf{82.44}
            & \textbf{91.87} & \textbf{84.96} \\
    
            4 bands
            & 95.93 & 92.18
            & 90.13 & 82.03
            & 91.73 & 84.72 \\
            \bottomrule
        \end{tabular}
        \par\vspace{3pt}
        \noindent\begin{minipage}{\linewidth}
        \PRcaptionfont
        The best results are highlighted in bold.
        \end{minipage}
    \end{table}

The three-band setting achieves the best overall results, whereas further increasing the partition to four bands leads to slight degradation. This suggests that insufficient partitioning limits frequency discrimination, while overly fine partitioning may fragment complementary frequency information and increase autoregressive modeling complexity. Therefore, three bands are adopted as the default configuration.

	

	\section{Conclusion}
	
	In this paper, we proposed Freq-RemoteVAR, a frequency autoregressive framework for remote sensing change detection. Different from conventional one-shot dense prediction methods, Freq-RemoteVAR reformulates change mask prediction as a next-frequency generation problem. To support this paradigm, we introduced a frequency-aware mask tokenization strategy to construct multi-frequency token targets from ground-truth change masks. We further designed a Frequency VAR Transformer conditioned on bi-temporal fused memory, enabling progressive change token generation under reliable pre/post-change image evidence. In addition, Scale-Aligned RoPE Cross Attention was developed to align frequency mask queries with bi-temporal condition tokens in a unified spatial coordinate system, while Change-quality Control adaptively modulates the generation process to suppress pseudo-change responses caused by shadows, illumination variations, texture inconsistency, and slight misregistration. Extensive experiments on CDD, GZ-CD, and LEVIR-CD demonstrate that Freq-RemoteVAR achieves superior quantitative performance and produces more complete change regions with sharper boundaries. These results verify the effectiveness of modeling remote sensing change detection from a low-to-high frequency autoregressive perspective.

	
	%

	

\section*{CRediT authorship contribution statement}

\noindent\textbf{Luqi Gong:} Conceptualization, Methodology, Software, Data curation, Formal analysis, Investigation, Validation, Visualization, Writing -- original draft, Writing -- review \& editing.
\textbf{Yue Chen:} Software, Validation, Visualization, Writing -- original draft.
\textbf{Qixin Xie:} Methodology, Software, Validation, Writing -- review \& editing.
\textbf{Ziqiang Chen:} Data curation, Formal analysis, Validation, Writing -- review \& editing.
\textbf{Chao Li:} Conceptualization, Supervision, Project administration, Writing -- review \& editing.
\textbf{Xuefeng Zhao:} Methodology, Supervision, Writing -- review \& editing.

\section*{Acknowledgements}

This research was supported by the Zhejiang Leading Innovative and Entrepreneurial Team (Chief Scientist) under Grant No. 2025R01002.

\section*{Data availability}

The CDD, GZ-CD, and LEVIR-CD datasets used in this study are publicly available from their original providers.

	\ifCLASSOPTIONcaptionsoff
	\newpage
	\fi
	
	\bibliographystyle{elsarticle-num}
	\bibliography{main}

\end{document}